\documentclass[letterpaper, 10 pt, conference]{ieeeconf}  

\IEEEoverridecommandlockouts                              

\usepackage{graphics} 
\usepackage[table,x11names]{xcolor}
\usepackage[acronyms]{glossaries}
\usepackage{booktabs} 
\usepackage{listings}
\usepackage{subcaption} 
\usepackage{arydshln}
\usepackage{nicematrix}
\usepackage{tikz}
\usetikzlibrary{decorations.markings, patterns}
\usepackage{pgfplots}
\pgfplotsset{compat=1.18}
\allowdisplaybreaks
\usepackage{amsmath,amssymb}
\usepackage{mathtools}
\makeatletter
\let\NAT@parse\undefined
\makeatother
\usepackage[hidelinks]{hyperref}
\usepackage{cite}

\usepackage[finalnew]{trackchanges} 
\addeditor{R10}
\addeditor{R13}
\addeditor{R14}

\allowdisplaybreaks

\makeatletter
\def\CheckMinus\ignorespaces{\@ifnextchar-{}{\phantom{-}}}
\makeatother
\newcolumntype{I}{>{\CheckMinus}c}
\newcolumntype{J}{>{\CheckMinus}l}
\newcolumntype{H}{>{\setbox0=\hbox\bgroup}c<{\egroup}@{}}

\newcommand{\xg}{x_g}
\newcommand{\xs}{x_s}
\newcommand{\Z}{\zeta}
\newcommand{\Zl}{\zeta_{l}}
\newcommand{\eye}{\mathbf{I}}

\newcommand{\nx}{n_x}
\def \nu {n_u}

\newcommand{\ncx}{n_{c_x}}
\newcommand{\ncu}{n_{c_u}}

\def \L {\mathcal{L}}
\newcommand{\xk}{x(k)}
\newcommand{\uk}{u(k)}
\newcommand{\xkp}{x(k+1)}
\newcommand{\pobj}{J}

\newcommand\norm[1]{\left\lVert#1\right\rVert}
\newcommand{\Qs}{Q}
\newcommand{\Rs}{R}

\newcommand{\NQ}{\Omega}

\newcommand{\lmdxk}{\lambda_{x}(k)}
\newcommand{\lmduk}{\lambda_{u}(k)}

\newcommand{\vardk}{\vartheta_{d}(k)}
\newcommand{\vardm}{\vartheta_{d}(k-1)}
\newcommand{\vars}{\vartheta_{s}}
\newcommand{\varg}{\vartheta_{g}}
\newcommand{\fxu}{f(\xk,\uk )}
\newcommand{\J}{\mathcal{J}}
\newcommand{\Zset}{\mathcal{Z}}
\newcommand{\inseg}{\Z^{\oplus}}
\newcommand{\acseg}{\Z^{\ominus}}
\newcommand{\inj}{j_{l}^{\oplus}}

\newcommand{\vpx}{\mathbf{v^p_x}} 
\newcommand{\vpu}{\mathbf{v^p_u}} 
\newcommand{\vsx}{\mathbf{v^s_x}} 
\newcommand{\vsu}{\mathbf{v^s_u}} 
\newcommand{\vdx}{\mathbf{v^d_x}} 
\newcommand{\vdu}{\mathbf{v^d_u}} 

\newcommand\xl[1]{x^l{\expandafter\ifx\expandafter\relax
  \detokenize{#1}\relax\else{(#1)}\fi}}
\newcommand\ull[1]{u^l{\expandafter\ifx\expandafter\relax
  \detokenize{#1}\relax\else{(#1)}\fi}}
  \newcommand\inJ[1]{\J_{#1}^{\oplus}}
\newcommand{\excsym}{\scalebox{0.45}{$\square$}}
\newcommand{\incsym}{\scalebox{0.45}{$\blacksquare$}}
\newcommand{\we}{w_{\excsym}}
\newcommand{\wi}{w_{\incsym}}
\newcommand{\De}{D_{\excsym}}
\newcommand{\Di}{D_{\incsym}}
\newcommand{\Fe}{F_{\excsym}}
\newcommand{\Fi}{F_{\incsym}}
\newcommand\win[1]{{w}_{\incsym}^{l}{\expandafter\ifx\expandafter\relax
  \detokenize{#1}\relax\else{(#1)}\fi}}
\newcommand\wex[1]{w_{\excsym}^{l}{\expandafter\ifx\expandafter\relax
  \detokenize{#1}\relax\else{(#1)}\fi}}
\newcommand\winb{{\overline{w}}_{\incsym}}
\newcommand\wexb{{\overline{w}}_{\excsym}}
\newcommand{\nw}{n_w}

\newcommand\pl[2]{p^l_{\expandafter\ifx\expandafter\relax
  \detokenize{#1}\relax\else{#1}\fi}{\expandafter\ifx\expandafter\relax
  \detokenize{#2}\relax\else{(#2)}\fi}}

\newcommand\xe[3]{x_{\expandafter\ifx\expandafter\relax
  \detokenize{#1}\relax\else{#1}\fi}^{\expandafter\ifx\expandafter\relax
  \detokenize{#2}\relax\else{#2}\fi}{\expandafter\ifx\expandafter\relax
  \detokenize{#3}\relax\else{(#3)}\fi}}
\newcommand\ue[3]{u_{\expandafter\ifx\expandafter\relax
  \detokenize{#1}\relax\else{#1}\fi}^{\expandafter\ifx\expandafter\relax
  \detokenize{#2}\relax\else{#2}\fi}{\expandafter\ifx\expandafter\relax
  \detokenize{#3}\relax\else{(#3)}\fi}}
\newcommand{\nb}{n_b}

\definecolor{gray2}{rgb}{0.4,0.4,0.4}%
\newcommand{\drawgt}{\raisebox{1pt}{
  \begin{tikzpicture}
    \draw[line width=0pt, fill=darkgray, fill opacity=0.25](2.mm,0) rectangle (3.5mm,1.5mm);
  \end{tikzpicture}%
}}

\newcommand{\drawpred}{\raisebox{1pt}{
  \begin{tikzpicture}
    \draw[dashed][color=red, dash pattern=on 0.25mm off 0.25mm](2.mm,0) rectangle (3.5mm,1.5mm);
  \end{tikzpicture}%
}}

\newcommand{\drawpredman}{\raisebox{1pt}{
  \begin{tikzpicture}
    \draw[dashed][color=red, dash pattern=on 0.25mm off 0.25mm, pattern=north west lines, pattern color=red](2.mm,0) rectangle (3.5mm,1.5mm);
  \end{tikzpicture}%
}}

\newcommand{\drawtraj}{\raisebox{2pt}{
\begin{tikzpicture}
    \draw[dotted, very thick][color=gray2, mark=o, mark options={solid, gray2, fill}](0,0) -- (3mm,0);
    \draw[color=gray2, mark=o, mark options={solid, gray2, fill}](0,0) -- (3mm,0);
\end{tikzpicture}
}}

\newcommand{\drawtraceQ}{\raisebox{2pt}{
\begin{tikzpicture}
    \draw[color=brown, line width=1.5pt](0,0) -- (3.5mm,0);
\end{tikzpicture}
}}

\newcommand{\drawgentraj}{\raisebox{2pt}{
\begin{tikzpicture}
    \draw[color=green, line width=1.5pt, mark=o, mark options={solid, green, fill}](0,0) -- (3.5mm,0);
\end{tikzpicture}
}}

\newcommand{\drawout}{\raisebox{2pt}{
  \begin{tikzpicture}
    \draw[blue,fill=blue] (0,0) circle (.25ex);
  \end{tikzpicture}%
}}

\newcommand{\drawQoutFP}{\raisebox{0pt}{
  \begin{tikzpicture}
    \draw[red, line width=1.0pt] (0,0) circle (.6ex);
  \end{tikzpicture}%
}}

\newcommand{\drawQoutTP}{\raisebox{0pt}{
  \begin{tikzpicture}
    \draw[green, line width=1.0pt] (0,0) circle (.6ex);
  \end{tikzpicture}%
}}

\def\ie{\textit{i.e.},}

\newcommand{\figref}[1]{\hyperref[#1]{Fig.~\ref*{#1}}}
\newcommand{\tabref}[1]{\hyperref[#1]{Tab.~\ref*{#1}}}
\newcommand{\secref}[1]{\hyperref[#1]{Sec.~\ref*{#1}}}

\newcommand{\figvspace}{\vspace{0em}}
\newacronym{lfd}{LfD}{Learning from Demonstration}
\newacronym{irl}{IRL}{Inverse Reinforcement Learning}
\newacronym{kkt}{KKT}{Karush–Kuhn–Tucker}
\newacronym{ikkt}{IKKT}{Inverse \gls{kkt}}
\newacronym{milp}{MILP}{Mixed-integer linear programming}
\newacronym{bnirl}{BN-IRL}{Bayesian Network Inverse Reinforcement Learning}
\newacronym{cbnirl}{CBN-IRL}{Constraint-based BN-IRL}
\newacronym{st}{s.t.}{such that}
\newacronym{gesd}{GESD}{Generalized Extreme Studentized Deviate}
\newacronym{pf}{p.f.}{Primal Feasibility}
\newacronym{df}{d.f.}{Dual Feasibility}
\newacronym{cs}{c.s.}{Complementary Slackness}
\newacronym{sc}{s.c.}{Stationary Condition}
\newacronym{ld}{l.d.}{Lagrange Derivative}

\title{\LARGE \bf
Jointly Learning Cost and Constraints from Demonstrations for Safe Trajectory 
Generation
}

\author{Shivam Chaubey, Francesco Verdoja, Ville Kyrki%
\thanks{This work was supported by Business Finland, decision 9249/31/2021.}%
\thanks{S. Chaubey, F. Verdoja, and V. Kyrki are with the School of Electrical 
Engineering, Aalto University, Espoo, Finland. 
{\tt\small \{firstname.lastname\}@aalto.fi}}%
}

\begin{document}
\maketitle
\thispagestyle{empty}
\pagestyle{empty}


\begin{abstract}
\Gls{lfd} allows robots to mimic human actions. However, these methods do not
model constraints crucial to ensure safety of the learned skill. Moreover, even
when explicitly modelling constraints, they rely on the assumption of a known
cost function, which limits their practical usability for task with unknown
cost. In this work we propose a two-step optimization process that allow to
estimate cost and constraints by decoupling the learning of cost functions from
the identification of unknown constraints within the demonstrated trajectories.
Initially, we identify the cost function by isolating the effect of constraints
on parts of the demonstrations. Subsequently, a constraint leaning method is
used to identify the unknown constraints. Our approach is validated both on
simulated trajectories and a real robotic manipulation task. Our experiments
show the impact that incorrect cost estimation has on the learned constraints
and illustrate how the proposed method is able to infer unknown constraints,
such as obstacles, from demonstrated trajectories without any initial knowledge
of the cost.
\end{abstract}


\glsresetall 

\section{Introduction}
\label{sec:intro}

Current robots are limited in their ability to perform complex tasks due to the
difficulty of programmatically describing the desired behaviors. The main
approach to address this challenge is \gls{lfd}, a method in which robots learn
to perform tasks effectively by imitating expert demonstrations, bypassing the
need for explicit programming of complex behaviors \cite{billard2008survey}.

However, learning generalizable policies with \gls{lfd} methods can be
challenging as these methods often do not explicitly guarantee constraint
satisfaction. This is a significant issue, since constraints such as workspace obstacles or velocity limits in part of the workspace
are used to represent essential safety requirements and performance specifications,
and without guarantees of constraint
satisfaction, the safety and efficacy of robotic systems cannot be ensured. 
Even outside \gls{lfd}, the importance of constraints is visible in recent
developments in \gls{irl} where work has been devoted to methods to prioritize
learning constraints over the traditional emphasis on reward functions.

One main limitation of existing methods for constraint learning is the
assumption that the reward function (or cost) is known, as observed in
\cite{gaurav2023learning, Malik, Scobee}. However, incorrect user-defined cost can cause constraints to be erroneously identified. Therefore, identifying both the cost
function and constraints simultaneously is necessary to learn skills while
ensuring safety of the system.

In this paper, we introduce a method to jointly learn both cost function and
constraints from human demonstrations. We achieve this with a novel method to
estimate cost under the assumption that constraints are affecting parts of the
demonstration and subsequently utilizing the retrieved cost to identify the constraints. The learned constraints can be applied to generate new trajectories. These trajectories not only align with the
observed human demonstrations but also prioritize safety, ensuring that the
robot's actions are both effective and secure. The main contributions of this
work are:
\begin{figure}
\begin{tikzpicture}
    \node[anchor=south west,inner sep=0] (image) at (0,0) {\includegraphics[width=\linewidth]{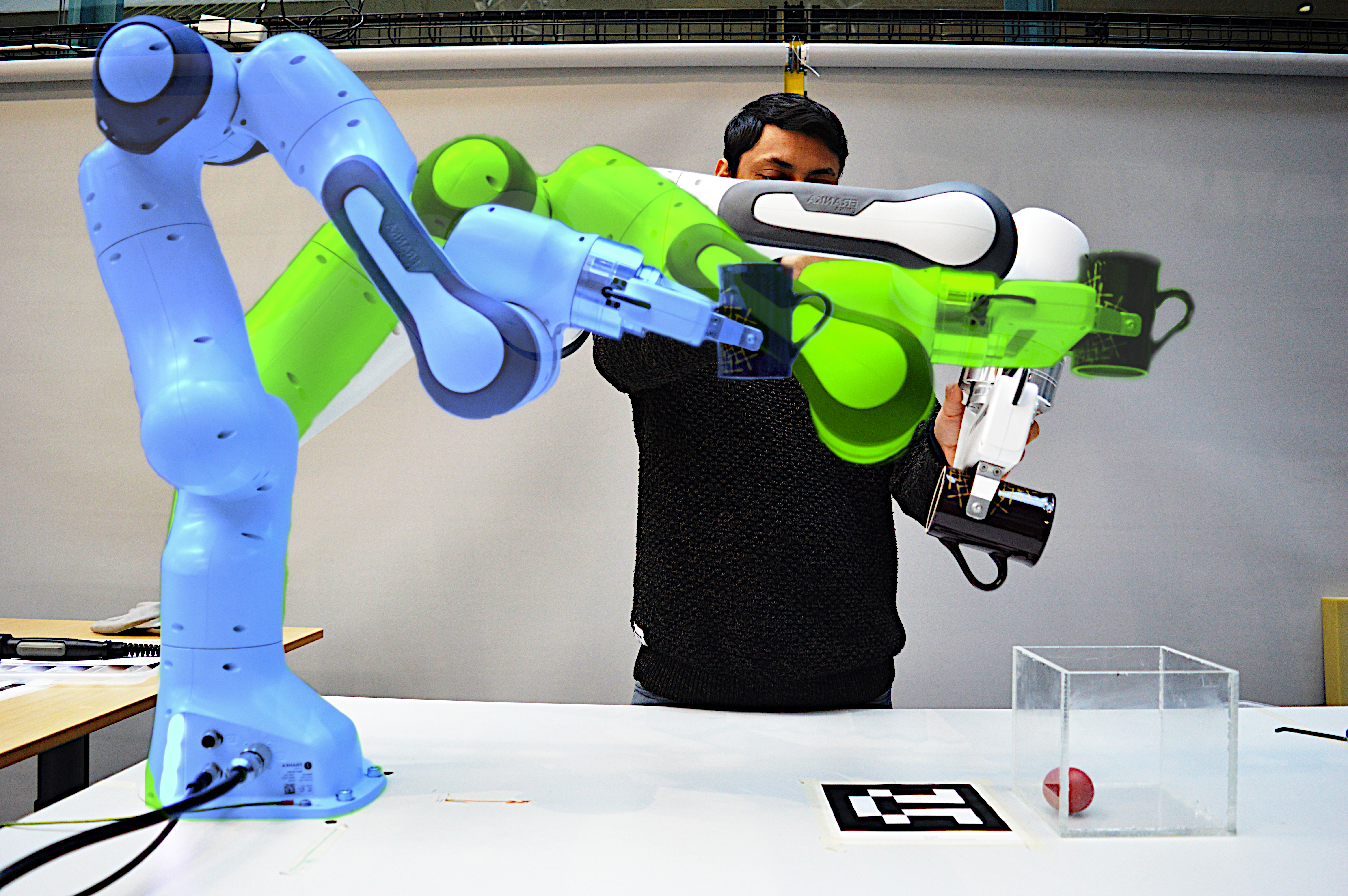}};
    \begin{scope}[shift={(0cm,0cm)}, scale=0.32] 
        \begin{scope}
            \fill[white] (-0.1,0.1) rectangle (9,8.0); 
        \end{scope}
        \begin{axis}[%
width=3.0in,
height=2.5in,
at={(0.544in,0.481in)},
scale only axis,
xmin=-0.238528302919752,
xmax=0.453259062110024,
xlabel style={font=\color{white!25!black}},
xlabel={$y$},
ymin=2.9,
ymax=5.0,
ylabel style={font=\color{white!25!black}},
ylabel={$\theta$},
axis background/.style={fill=white},
title style={font=\bfseries},
axis x line*=bottom,
axis y line*=left,
label style={font=\Large, fill=white},
tick label style={font=\Large},  
]
\addplot [color=gray2, mark=*, line width=3.0pt, mark options={solid, gray2, fill}]
  table[row sep=crcr]{%
-0.162304969042727	4.72825553367375\\
-0.138544172225028	4.74201038102013\\
-0.0873745616061391	4.72585392402912\\
-0.0107106419915068	4.71564309960082\\
0.0757192196340581	4.71649889752267\\
0.168683528256735	4.71112583973942\\
0.254234896601672	4.69705095569395\\
0.312357775597946	4.70410143151001\\
0.350499961032653	4.71437262622607\\
0.382383599317027	4.71990738196488\\
0.398998816000186	4.72462054212897\\
0.403259062110024	4.73309055429854\\
0.402824265512617	4.73203125422856\\
0.396260719018818	4.60059299099816\\
0.382935376754645	4.36000363415357\\
0.373004437441763	4.13383897076503\\
0.371701257972357	3.97787272029\\
0.371660127666125	3.84581348460271\\
0.370915838455131	3.71775498487446\\
0.371270808524335	3.58900370036244\\
0.36890854647649	3.46913005259185\\
0.368679091575567	3.39646287271181\\
0.368944026849451	3.39343617615708\\
0.369467040922976	3.39553336275077\\
0.365621901182212	3.36616696054891\\
0.362997390804024	3.33601792393516\\
0.355530150740557	3.30360044427809\\
0.351505011504989	3.27396966956699\\
0.347442899678719	3.23622782168494\\
0.34213383120673	3.18990702494071\\
};
\addplot[area legend, line width=2.5pt, dashed, draw=red, pattern=north west lines, pattern color=red]
table[row sep=crcr] {%
x	y\\
-9.70221251947686	-5.38085896789839\\
-9.70221251947686	4.7041014315100\\
0.34213383120673	4.7041014315100\\
0.34213383120673	-5.38085896789839\\
}--cycle;

\end{axis}
    \end{scope}

\end{tikzpicture}
\caption{Starting from a human demonstration, we propose a novel method to
jointly learn both cost and constraints to allow the robot to replicate the task. \add{Here, a user demonstrates dropping a ball from a cup to a target; the retrieved constraint limits the cup tilt ($\theta$) when the cup position ($y$) is not over target, as shown in red in the graph.}}
\label{fig:cover}
\figvspace{}
\end{figure}

\begin{itemize}
\item A method jointly learning optimal cost and constraints for skills
generated through human demonstrations in the presence of unknown constraints
and cost.
\item An experimental validation of our approach both in simulation and with a
real-world system.
\item Code for our method and experiments to facilitate reproducibility and
further research in this area\footnote{Source code for method and experimental setting of this paper is found at \href{https://version.aalto.fi/gitlab/chaubes1/jointly-learning-cost-and-constraints}{version.aalto.fi/gitlab/chaubes1/jointly-learning-cost-and-constraints}.}.
\end{itemize}

Our paper is structured as follows: In \secref{sec:related_work}, we discuss
recent developments in the field of task learning and safety assurance,
examining related work, and providing context for our approach.
\secref{sec:problem} presents the problem formulation addressed by our method.
\secref{sec:Method}, we introduce our proposed method for extracting cost
functions and constraints, describing the techniques and algorithms used.
Finally, \secref{sec:Experiment} \change{conducts a comprehensive}{presents an} evaluation of our
approach, including its application in real-world scenarios, and conclusions are
drawn in \secref{sec:conclusions}.

\section{Related Work}
\label{sec:related_work}

Recent studies \cite{ding2022x, fischer2021sampling,
NEURIPS2020_a4c42bfd}, have identified three main categories within \gls{lfd}:
policy, plan, and cost or reward learning, based on the learning outcomes.
Although these approaches offer powerful ways to train agents, they often
struggle to ensure that important constraints are met
\cite{ravichandar_recent_2020}.

Most of the methods proposed to infer constraints require prior knowledge of the
cost function \cite{gaurav2023learning, Malik, Scobee, chou2020learning1, park2020inferring}. The method proposed in \cite{Malik, Scobee} represents an
initial step in learning arbitrary constraints. However, it assumes the reward
function, which is required to calculate the policy, is known or calculated
using an \gls{irl} method. In another approach \cite{chou2020learning1},
constraints are inferred by comparing demonstrated trajectories with those of
higher performance from a nominal model. These constraints are found within the
feature space of lower-cost trajectories not followed by the demonstrator,
implying the presence of previously unidentified constraints. However, this
approach relies on an already known cost function and lacks a mechanism for
determining an optimal cost function.

The method proposed in \cite{gaurav2023learning} has limitations, including the
extraction of only a single constraint and
exclusive focus on soft constraints, which may not be suitable for
safety-critical applications. The work by \cite{park2020inferring} recovers
constraints provided reward function by introducing  \gls{bnirl} and its
extension, \gls{cbnirl}. While \gls{bnirl} organizes decision-making without
strict constraints, \gls{cbnirl} adds hard constraints but struggles with
accurately identifying constrained tasks due to its flexible assignment
variable. In the approach presented by \cite{chou2020learning}, constraints are
primarily extracted by applying \gls{kkt} conditions, considering parametric
uncertainty in cost up to a certain degree to handle locally-optimal
demonstrations.

Our method contributes to the field of cost and constraint learning, addressing
the limitations of previous approaches. Unlike prior methods that concentrate
solely on soft constraints or require prior knowledge of the cost function,
our approach segments demonstrations to isolate the impact of constraints on
policy learning. This segmentation facilitates a more nuanced understanding of
how policy influences decision-making without constraints, thereby allowing for
a more accurate and robust inference of constraints directly from the learned
policy and the demonstrations themselves.

\section{Problem formulation}
\label{sec:problem}

\subsection{Optimal trajectory problem}
\label{sec:opti_traj}

Let us consider a set of optimal human demonstrations $\Zset = (\zeta_1, \dots,
\zeta_L)$, where each trajectory $\zeta_l = (x^l, u^l)$ consists of a sequence
of states $x^l$ and controls $u^l$. Within each trajectory, individual points in
time are denoted $(\xl{k}, \ull{k})$. For simplicity, we momentarily omit the
superscript $l$. 

Each demonstration is assumed to be the solution to constrained quadratic
optimization problem 
\begin{subequations}
\begin{equation} \label{eq:objective_}
\min_{\xk, \uk } \pobj 
\end{equation}
\gls{st}
\begin{align} \label{eq:eq_cons}
\underbrace{
\begin{array}{l}
\xkp  - f(\xk, \uk ) = 0\\
\xk  - \xs = 0\\
\xk  - \xg = 0
\end{array}
}_{\text{Equality constraints}}
\begin{array}{l}
\begin{aligned}    
\forall k& \in [1, N -1]\\
 k& = 1 \\
k&=N
\end{aligned}
\end{array}
\end{align} 
\begin{align} \label{eq:ineq_cons}
\underbrace{
\begin{array}{l}
g(\xk) \leq 0 \\
h(\uk ) \leq 0
\end{array}
}_{\text{Inequality constraints}}
\quad
\begin{array}{r}
\begin{aligned}    
\forall k& \in [1, N]\\
\forall k& \in [1, N-1]
\end{aligned}
\end{array}
\end{align}
\end{subequations}
where $\pobj$ is the standard quadratic cost with parameters $Q$ and $R$,

\begin{equation} \label{eq:initial_problem}
\pobj = \frac{1}{2} \left(\sum_{k = 1}^{N}  \xk ^{\top} Q \xk  +  
\sum_{k = 1}^{N-1} \uk  ^{\top} R \uk  \right)\enspace,
\end{equation}
and $\xk \in \mathbb{R}^{\nx}$ represents the state vector, while $\uk  \in \mathbb{R}^{\nu}$ denotes the control vector. The parameter $Q$ serves as the weight for penalizing states, contributing to the generation of the optimal trajectory. On the other hand, $R$ is the weight applied to minimize control efforts. The inequality constraint, represented by $g(\xk): \mathbb{R}^{\nx} \to \mathbb{R}^{\ncx}$ for the states and $h(\uk ): \mathbb{R}^{\nu} \to \mathbb{R}^{\ncu}$ for the control, encloses all the known/unknown inequality constraints, respectively. Here, $\ncx$ and $\ncu$ denote the number of inequality constraints for states and control, respectively. The constraint $\xkp  - f(\xk, \uk ) = 0$ represents known system dynamics. Lastly, conditions $x(1) - \xs = 0$ and $x(N) - \xg = 0$ correspond to the initial and final states, respectively.
\subsection{Constraints formulation}
\label{sec:constr_form}

The constraint set defined in \eqref{eq:eq_cons} and \eqref{eq:ineq_cons}
includes both equality and inequality constraints. The equality constraints are
assumed to be known and account for the system dynamics governed by
$f(\xk,\uk)$, and initial and final states, \ie{} $\xs$ and $\xg$. These
constraints ensure that the system starts and ends in specified states and
adheres to the predefined dynamics throughout its operation. The inequality
constraints for state $g(\xk)$ and control $h(\uk)$ can be known or unknown.
These constraints limit the states and controls to prevent them from exceeding
certain bounds, thereby ensuring safety and feasibility within the system's
operational environment. The constraints are considered to be vector valued, and
thus, an arbitrary number of constraints can be used.

Moreover, while not explicitly specified in \eqref{eq:ineq_cons}, inequality
constraints can be divided into inclusive (\scalebox{0.6}{$\blacksquare$}) and
exclusive (\scalebox{0.6}{$\square$}). Inclusive constraints ensure that the
trajectory remains within specified bounds, \change{catering to}{promoting} safety and task
compliance, while exclusive constraints are designed to keep the trajectory
outside of certain bounds, thus avoiding unsafe regions. We also assume inequality
constraints to be describable as axis-aligned bounds for inclusive constraints
and axis-aligned shapes for exclusive constraints, noting that axis-aligned
constraints can represent irregular shapes when utilized in large numbers. In the following, the vector $w(k) \in \mathbb{R}^{\nw}$ may represent either the state vector $\xk \in \mathbb{R}^{\nx}$ or the action vector $\uk\in \mathbb{R}^{\nu}$, depending on the problem context. 

Given lower-bounds $w_{\incsym}^{lb}, w_{\excsym}^{lb} \in \mathbb{R}^{\nw}$ and upper-bounds $w_{\incsym}^{ub}, w_{\excsym}^{ub} \in \mathbb{R}^{\nw}$, inclusive constraints
can be expressed as
\begin{equation}\label{eq:incl_cons}
w_{\incsym}^{lb} \leq \wi(k) \leq w_{\incsym}^{ub} \quad 
\forall k \in [1, N]\enspace,
\end{equation}
while exclusive constraints can be expressed as
\begin{multline}\label{eq:excl_cons}
    \bigvee_{i=1}^{\nw}
    \we(k,i) - w_{\excsym}^{lb}(i) \leq 0 \lor 
    \we(k,i) + w_{\excsym}^{ub}(i) \leq 0 \\
    \forall k \in [1, N]~,
\end{multline}
where $i \in [1, \nw]$ refers to the $i$-th dimension out of the $\nw$ state or
action dimensions. The interpretation of \eqref{eq:excl_cons} is that, in order
for the demonstration to not violate an exclusive constraint, each time-step $k$
needs to be outside the constraint's bounds on at least one dimension.

Please note that, while inclusive constraints are convex, exclusive constraints
are not, and therefore they need to be relaxed when included in convex
optimization. For this reason, in this work, while we present a method able to retrieve unknown constraints of both types, in the experiments we focus on exclusive constraints only. 

\subsection{Cost and constraints estimation problem}
\label{sec:extract_problem}

In the setting of this paper, some of the inequality constrains as well as the
cost parameters are unknown. A significant challenge arises from the simultaneous determination of the cost function and multiple constraints, leading to the issue of unidentifiability: numerous configurations of cost and constraint functions may produce identical trajectories \cite{ng2000algorithms}.

Therefore, in this work we propose a method to jointly identify both cost
parameters ($Q$ and $R$) as well as an arbitrary number of unknown inequality
constraints affecting the demonstrations. Our method also addresses the
relaxation of exclusive constraints.

\section{Method}
\label{sec:Method}

To jointly estimate both cost parameters and constraints, our approach is
twofold: first, we determine the cost associated with the demonstration under
the assumption that the unknown constraints are active only at certain time
steps. This step isolates the effect of the constraints on the cost function and
captures the underlying cost structure that drives the system's behavior in the
absence of external unknown constraints.
Second, we infer the unknown constraints based on the previously estimated cost. This step
relies on the understanding that deviations of the trajectory from the
unconstrained behavior can be attributed to the unknown constraints. 

We will describe our approach for estimating the cost parameters in
\secref{sec:cost_ext}, followed by the estimation of the constraints in
\secref{sec:cons_ext}. But before, in \secref{sec:kkt_cond} we present a
reformulation of the trajectory optimization problem presented in
\secref{sec:opti_traj} according to the \gls{kkt} conditions which is used in
multiple parts of our method.

\subsection{KKT formulation}
\label{sec:kkt_cond}

\change{We assume that the}{The} optimization problem presented in \secref{eq:initial_problem} is convex, differentiable, and satisfies constraint qualification criteria \cite{ scholkopf2002learning}. \change{These assumptions}{This} enables us to utilize the \add{first order} \gls{kkt} conditions \cite{karush1939minima, kuhn2013nonlinear} to formulate optimization problems within a unified framework, capable of handling both equality and inequality constraints alongside the objective function.
Given our particular interest in the inverse problem of \ref{eq:initial_problem}, where solutions (trajectory $\Zl$) are known and unknown costs and constraints need to be learned, we will adapt the \gls{kkt} conditions, referred to as the \gls{ikkt} conditions, to infer these unknowns.

In the \gls{ikkt} formulation, our goal is to derive unknown costs and constraints from given demonstration while considering some known constraints. We introduce specific vectors within our \gls{ikkt} framework that represent various settings for extracting costs and constraints. Specifically, we have binary switching vectors for state $(\vpx, \vdx, \vsx)$ and control $(\vpu, \vdu, \vsu)$ for different conditions. In these vectors, an element of `1' signifies that the dimension is included in the optimization formulation, while a `0' means it is excluded. We define $\vdx = [\vsx, \vsx]^{\top}$ and $\vdu = [\vsu, \vsu]^\top$. 

This approach in the \gls{ikkt} formulation allows us to learn about costs and constraints without needing to alter the formulation. We reinterpret the \gls{kkt} stationary condition; rather than setting it to zero, we represent this condition with a Lagrange derivative expression that needs to be minimized. This modification allows us to address cases where achieving the exact condition might not be possible or optimal, focusing on minimizing the deviation from the ideal stationary state. The \gls{ikkt} formulation of our problem is then expressed as follows:\\
    \gls{pf}:    
        \begin{equation}
        \begin{gathered}
        \vpx \odot g(\xk) \leq 0, \ 
        \vpu \odot h(\uk) \leq 0
        \end{gathered}
            \label{eq:primal_feas_out}
        \end{equation}
    \gls{df}:    
        \begin{equation}
        \begin{gathered}
        \vdx \odot \lmdxk \geq 0, \ 
        \vdu \odot \lmduk \geq 0
        \end{gathered}
            \label{eq:dual_feas_out}
        \end{equation}
    \gls{cs}:
            \begin{equation}
            \begin{gathered}
            \vdx \odot \lmdxk \odot g(\xk) = 0, \\
            \vdu \odot \lmduk \odot h(\uk ) = 0 
            \end{gathered}
                \label{eq:comp_slack_out}
            \end{equation}
    \gls{ld}: 
\allowdisplaybreaks{
    \begin{align}
           &\left.\frac{\partial \L}{\partial \xk}\right|_{k = 1} = 
           Q\xk + \vsx \odot \left[\lmdxk^{\top} \frac{\partial g(\xk)}{\partial \xk}\right]^{\top} \nonumber \\ 
           &  \qquad \qquad \qquad  - \left[\vardk^{\top} \frac{\partial \fxu}{\partial \xk}\right]^{\top} + \vars  \nonumber\\ 
           &\left.\frac{\partial \L}{\partial \uk }\right|_{k = 1} = 
            R\uk  +  \vsu \odot \left[ \lmduk^{\top} \frac{\partial h(\uk )}{\partial \uk } \right]^{\top} \nonumber \\
            & \qquad \qquad \qquad \qquad   - \left[\vardk^{\top} \frac{\partial \fxu}{\partial \uk }\right]^{\top} \nonumber \\ 
            \cdashline{1-2}
           &\left.\frac{\partial \L}{\partial \xk}\right|_{k \in (1, N)} = 
           Q\xk + \vsx \odot \left[\lmdxk^{\top} \frac{\partial g(\xk)}{\partial \xk}\right]^{\top} \nonumber\\
           &\qquad \qquad \  + \vardm - \left[\vardk^{\top} \frac{\partial \fxu}{\partial \xk}\right]^{\top} \\ 
           &\left.\frac{\partial \L}{\partial \uk }\right|_{k \in (1, N)} =  
            R\uk  +  \vsu \odot \left[ \lmduk^{\top} \frac{\partial h(\uk )}{\partial \uk } \right]^{\top} \nonumber \\
            &\qquad \qquad \qquad \qquad  - \left[\vardk^{\top} \frac{\partial \fxu}{\partial \uk }\right]^{\top} \nonumber\\ 
            \cdashline{1-2}
           &\left.\frac{\partial \L}{\partial \xk}\right|_{k = N} = Q\xk + \vsx \odot \left[\lmdxk^{\top} \frac{\partial g(\xk)}{\partial \xk}\right]^{\top} \nonumber \\  
           & \qquad \qquad \qquad \qquad \qquad + \vardm + \varg \nonumber 
        \label{eq:stat_con_out}
    \end{align}}

 Here, $\lmdxk \in \mathbb{R}_{\geq 0}^{\ncx}$ and $\lmduk \in \mathbb{R}_{\geq 0}^{\ncu}$  represents the vector of Lagrange multipliers associated with the inequality constraints in state and control respectively. $\vardk \in \mathbb{R}^{\nx}, k \in [1, N]$ denotes the vector of Lagrange multipliers linked to system dynamics. $\vars \in \mathbb{R}^{\nx}$ and $\varg \in \mathbb{R}^{\nx}$ denotes the vector of Lagrange multipliers linked to initial point $k = 1$ and final point $k=N$, respectively.

\subsection{Cost extraction} 
\label{sec:cost_ext}
We propose a two-step method to determine the cost parameters: initially,
demonstrations are locally partitioned into segments based on whether the
unknown constraints are active; subsequently, the inactive segments are utilized
to learn the cost. 

In cost learning setup, we set the vectors $\vpx$ and $\vpu$ to zeros. This step is taken to overlook the primal feasibility condition, as the demonstrations already meet this criterion for both known and unknown constraints. To understand the policy in the absence of unknown constraints, we introduce the vectors $\vsx$ and $\vsu$. The elements within the vectors $\vsx$ and $\vsu$ are set to `0' for unknown constraints and to `1' for known constraints. 

\subsubsection{Demonstration segmentation}
\label{sec:demo_segm}

We want to partition a trajectory $\Z$ into active segments $\acseg \subset
\Z$---\ie{} where the unknown constraints are active---and inactive segments
$\inseg \subset \Z$---\ie{} where the unknown constraints are inactive. Let us
define the three-point segment of trajectory. Three points were chosen as the
minimum required for solving the discrete \gls{kkt} formulation. $\Z$ starting at time-step $j$ as
$\Z^j = \{\Z(j), \Z(j+1), \Z(j+2)\}$. We propose a method to assign each
segment $\Z^j$ to either $\acseg$ or $\inseg$ leveraging the intuition that
inactive segments will behave according to the optimal cost parameters and known
constraints, while active segment---due to influence of unkown constraints---will
showcase a noticeable variation in the trace of the normalized cost $\NQ^j$,
obtained by normalizing $\Qs^j$ with regards to $\Rs^j$. Therefore, we propose
to estimate cost parameter for each segment $\Z^j$ with $j \in [1, N-2]$ and
then perform an outlier detection step to identify the active indices. 

Therefore, given a trajectory $\Z$ and an index $j$, we solve
\begin{equation}\label{eq:outlier_detection}
    \min_{
    \substack{\Qs^{j}, \Rs^{j}, \hat{\lambda}^j_{x}, \hat{\lambda}^j_{u}\\
    \hat{\vartheta}^j_{d}, \hat{\vars}^j, \hat{\varg}^j}}
    \norm{\J^j}_{2}
\end{equation}
\gls{st} \eqref{eq:dual_feas_out} and \eqref{eq:comp_slack_out} satisfy,
$\Qs^j \geq 0$, $\text{Tr}(\Qs^j) = 1$ and $\Rs^j > 0$, and
where
\begin{align*}
    \J^j &= \left[
        \frac{\partial \L}{\partial x(j)}, 
        \frac{\partial \L}{\partial u(j)},
        \frac{\partial \L}{\partial x(j+1)}, 
        \frac{\partial \L}{\partial u(j+1)}, 
        \frac{\partial \L}{\partial x(j+2)}
    \right]^{\top}\\
    \hat{\lambda}^j_{x} &=
        [\lambda_{x}(j), \lambda_{x}(j+1), \lambda_{x}(j+2)]\\
    \hat{\lambda}^j_{u} &=
        [\lambda_{u}(j), \lambda_{u}(j+1)]\\
    \hat{\vartheta}^j_{d} &=
        [\vartheta_{d}(j), \vartheta_{d}(j+1)]\\
    \hat{\vars}^j &= 0 \Leftarrow j \neq 1, \quad 
    \hat{\varg}^j = 0 \Leftarrow j \neq N-2\enspace.
\end{align*}

Equation \eqref{eq:outlier_detection} gives $\Qs^{j}$, $\Rs^{j}$, \ie{} an
estimate of the cost parameters for that segment. To detect outliers, the
normalized cost $\text{Tr}(\NQ^{j})$ is calculated for each segment at
time-step $j$ and the \gls{gesd} \cite{gesd} outlier procedure is performed on
list of $\text{Tr}(\NQ^j) \forall j \in [1, N]$. We then include all
$\Z^j$ identified as outlier into $\acseg$ and set $\inseg = \Z \setminus
\acseg$.

\subsubsection{Cost estimation from inactive segments}

To determine the unknown cost weights $Q$ and $R$, we consider all indices $\inj$ of inactive
segments $\inseg_l$ within each trajectory $\Zl$. This approach addresses the
issue of unidentifiability by focusing exclusively on time steps that fully
inform the demonstrated policy. It is important to note that unknown constraints
only prompt behavioral changes when active. To learn a time-invariant cost that
aligns with the entire set of demonstrations $\Zset$, we solve
{\begin{equation} \label{eq:cost_extraction}
        \min_{\substack{\Qs, \Rs, \tilde{\lambda}_{x}, \tilde{\lambda}_{u}\\ \tilde{\vartheta}_{d}, \tilde{\vars}, \tilde{\varg}}} \norm{[\inJ{1}, \inJ{2}, \ldots, \inJ{L}]^{\top}}_{2}
\end{equation}
\gls{st} \eqref{eq:dual_feas_out} and \eqref{eq:comp_slack_out} satisfy, $\Qs^j \geq 0$, $\text{Tr}(\Qs) = 1$ and $\Rs^j > 0$}, and
where
\begin{align*}
    &\inJ{l} = \left[\frac{\partial \L_l}{\partial x(q_l)}, \frac{\partial \L_l}{\partial u(q_l)}, \ldots \right]\\
    &\tilde{\lambda}_{x} = [ \lambda^l_x(q_l), \ldots], \  \tilde{\lambda}_{u} = [ \lambda^l_u(q_l), \ldots],\ \tilde{\vartheta}_{d} = [\vartheta^l_{d}(q_l), \ldots] \\
    &\tilde{\vars} = [ \vars^l(q_l), \ldots], \ \tilde{\vars}^l(q_l) = 0 \Leftarrow q_l \neq 1\\
    & \tilde{\varg} = [ \varg^l(q_l), \ldots],\ \tilde{\varg}^l(q_l) = 0 \Leftarrow q_l \neq N_l\\
    & \forall q_l \in \inj
\end{align*}

This optimization problem is solved once, incorporating all inactive segments
from all demonstrated trajectories to estimate the cost weights $\Qs$ and $\Rs$. 
\add{The computational cost of the problem is similar to that of a least squares problem.}
In the next section, we use these cost weights to identify the unknown
constraints.

\subsection{Constraint extraction} 
\label{sec:cons_ext}
Our proposed method for extracting constraints from observed demonstrations comprises three essential steps: Firstly, we start with the formulation of inequality constraints. Secondly, we introduce additional complementary conditions to effectively manage both inclusive constraints and unobserved bounds. Finally, we present a constraint learning algorithm that utilizes the initial two steps.

\subsubsection{Inequality constraints representation} \label{sec:ex_cons}

As detailed in \secref{sec:constr_form}, inequality constraints can be either inclusive or exclusive. We represent inclusive constraints as
\begin{equation}
\Di \wi(k) - \Fi \odot \winb \leq 0  
\label{eq:inc_cons}
\end{equation}
where
\begin{equation*}
\Di =
\begin{pmatrix*}[J] 
-\eye_{\nw \times \nw} \\
 \eye_{\nw \times \nw}
\end{pmatrix*}, \ 
\Fi =
\begin{pmatrix*}[J]
-\mathbf{1}_{\nw \times 1}\\
\mathbf{1}_{\nw \times 1}
\end{pmatrix*}, \ 
\winb =
\begin{pmatrix*}
w_{\incsym}^{lb} \\
w_{\incsym}^{ub}
\end{pmatrix*}.
\end{equation*}

Exclusive constraints are non-convex and need to be relaxed when included in convex optimization. 
To relax the formulation presented in \eqref{eq:excl_cons} into a convex set, we follow the method proposed by Schouwenaars \textit{et al.} \cite{schouwenaars2001mixed}, please refer to that work for a complete explanation. Let us introduce a binary
variable vector $p(k) \in \{0,1\}^{2\nw}$.
Each exclusive constraint can be represented as:
\begin{equation} \label{eq:exc_cons}
\De \we(k) - \Fe \odot \wexb - M(1 - p(k)) \leq 0 
\end{equation}
and
\begin{equation} \label{eq:exc_p_cons}
\|p(k)\|_1 \geq 1
\end{equation}
where $M$ denotes a large positive number and
\begin{equation*}
\De = 
\begin{pmatrix*}[J] 
\eye_{\nw \times \nw} \\
-\eye_{\nw \times \nw}
\end{pmatrix*}, \ 
\Fe =
\begin{pmatrix*}[J]
\mathbf{1}_{\nw \times 1} \\
-\mathbf{1}_{\nw \times 1}
\end{pmatrix*}, \ 
\wexb =
\begin{pmatrix*}
\we^{lb} \\
\we^{ub}
\end{pmatrix*}.
\end{equation*} 

Equation \eqref{eq:exc_cons} ensures compliance with the original constraints, distinguishing between active and inactive constraints, while \eqref{eq:exc_p_cons} ensures that at least one constraint is always active. This relaxation allow exclusive constraints to be used in an \gls{milp} optimization problem whose sub-problems are convex. 

\subsubsection{Complementary exclusive constraints}
We introduce two additional complementary exclusive conditions. The first condition helps to identify bounds that remain inactive throughout the demonstration due to the trajectory discretization. This is a known problem with discrete state-spaces not solved in previous works \cite{chou2020learning}. This complementary constraint guarantees that any unknown inactive bound, orthogonal to an active constraint, is positioned at or beyond the active point orthogonally. This effectively defines the limits of movement in directions not explicitly restricted by the observed demonstration. For each exclusive constraint, let us define 
\begin{equation} \label{eq:exc_comp_e}
\begin{pmatrix*}
e_{\excsym}^{lb} \\
e_{\excsym}^{ub}
\end{pmatrix*} = \De \we(k) - \Fe \odot \wexb\enspace,
\end{equation}
then, we construct the following complementary constraint:
\begin{align}    
\begin{aligned}    
    &\lambda^{lb}_{\excsym}(r)  e^{lb}_{\excsym}(t) \geq \epsilon ,  \lambda^{lb}_{\excsym}(r)  e^{ub}_{\excsym}(t) \geq \epsilon, \\
    &\lambda^{ub}_{\excsym}(r)  e^{lb}_{\excsym}(t) \geq \epsilon ,  \lambda^{ub}_{\excsym}(r)  e^{ub}_{\excsym}(t) \geq \epsilon \\
    & \forall r, t \in [1, \nw], r \neq t
\label{eq:met_comp_exc1}
\end{aligned}
\end{align}

In these equations, $\lambda^{lb}_{\excsym}(r), \lambda^{ub}_{\excsym}(r) \subset \lambda_w$ denote Lagrange multipliers for the lower and upper bounds of the point $\we$ in dimension $r$, respectively, and $\lambda_w$ is either $\lambda_x$ or $\lambda_u$ depending on the context. Similarly, $e^{lb}_{\excsym}(t)$ and $e^{ub}_{\excsym}(t)$ represent the lower and upper bound constraints for dimension $t$ as defined in \eqref{eq:exc_comp_e}. The equations assert that for all distinct pairs $(r, t)$, the product of each Lagrange multiplier with its orthogonal constraint must be greater than or equal to a small negative value $\epsilon$. This formulation ensures that if any bound in a dimension  is active for a point $\we(k)$ within the trajectory, non-active bounds in orthogonal dimensions remain within a reasonable margin from that point. 

A second additional constraint ensures that Lagrange multipliers are nullified for constraints that are inactive, effectively isolating non-active constraints from the \gls{ld} expression, \gls{df}, and \gls{cs} conditions. Formally, this idea is captured as:
\begin{equation} \label{eq:met_comp_exc2}
    \ (1 - p_r(k)) \odot \lambda_r(k) = 0\enspace. 
\end{equation}

The above formulation mandates that for a given dimension $r$ and at time-step $k$, the value of the Lagrange multiplier $\lambda_r(k) \subset \lambda_w$ must be zero if the associated constraints are inactive or non-binding---\ie{} $p_r(k) = 0$. 

\subsubsection{Constraint learning formulation}
Given the estimated cost weights $\Qs$ and $\Rs$ according to \secref{sec:cost_ext}, we identify the set of $\nb$\add{\footnote{The value of $\nb$ can be set higher if the number of unknown constraints is not known, resulting in overlapping constraints or constraints lying within others. The union of these constraints will provide the final constraints.}} unknown constraints acting on the
demonstrations\remove{ as follows}. Unlike to the cost learning approach outlined in Equation \eqref{eq:cost_extraction}, which omits unknown constraints, our focus here shifts towards learning those constraints. Therefore, to ensure \gls{pf}, we assign a `1' to the elements of vectors $\vpx$ and $\vpu$ that correspond to unknown constraints, and a `0' for those representing known constraints. We set every element within the vectors $\vsx$, $\vsu$, $\vdx$, and $\vdu$ to one. The revised formulation is as follows:  
\begin{equation} \label{eq:cons_extraction}
    \min_{\substack{\tilde{\lambda}_{x,k}, \tilde{\lambda}_{u,k}\\ \tilde{\vartheta}_{d,k}, \tilde{\vars}, \tilde{\varg}}} \norm{[\J_{1}, \J_{2}, \ldots, \J_{L}]^{\top}}_{2}
\end{equation}
\gls{st} \eqref{eq:primal_feas_out}, \eqref{eq:dual_feas_out}, \eqref{eq:comp_slack_out},  \eqref{eq:exc_p_cons}, \eqref{eq:met_comp_exc1}, and \eqref{eq:met_comp_exc2} satisfy, and where
\begin{align*}   
    \begin{aligned}
        &\J_{l} = \left[\frac{\partial \L_l}{\partial \xl{k}}, \frac{\partial \L_l}{\partial \ull{k}}, \ldots \right] , \\
        &\tilde{\lambda}_{x} = [ \lambda^l_{x}(k), \ldots], \
        \tilde{\lambda}_{u} = [ \lambda^l_{u}(k), \ldots],\
        \tilde{\vartheta}_{d} = [\vartheta^l_{d}(k), \ldots], \\
        &\tilde{\vars} = [ \vars^l(k), \ldots], \ \tilde{\vars}^l = 0 \Leftarrow k \neq 0 \\
        & \tilde{\varg} = [ \varg^l(k), \ldots],\ \tilde{\varg}^l = 0 \Leftarrow k \neq N_l\\
        & \forall k \in [1, N_l], \forall l \in [1, L] 
    \end{aligned}
\end{align*}

The above formulation includes inequality constraints for both states, denoted by $g(\xk)$, and controls, represented as $h(\uk)$. These constraints are detailed according to the definitions provided in equations \eqref{eq:inc_cons} and \eqref{eq:exc_cons}.

The bilinear variable terms in the model are simplified using the Matlab tools Yalmip \cite{Lofberg2004}. Following this, problems \eqref{eq:outlier_detection} and \eqref{eq:cost_extraction} are solved using semidefinite solvers, while the problem \eqref{eq:cons_extraction} is addressed using \gls{milp} optimization solvers.
\section{Experiments}
\label{sec:Experiment} 

\begin{figure*}
    \centering
    \begin{subfigure}[t]{0.24\textwidth}
%
\definecolor{mycolor1}{rgb}{0.90000,0.27000,0.27000}%
\definecolor{mycolor2}{rgb}{0.27000,0.90000,0.90000}%
\definecolor{gray2}{rgb}{0.4,0.4,0.4}%
\definecolor{black2}{rgb}{0.0,0.0,0.0}%

\begin{tikzpicture}[scale=0.35]

\begin{axis}[%
width=4.5in,
height=4.5in,
at={(0.751in,1.253in)},
scale only axis,
xmin=5.13292433537833,
xmax=74.8670756646217,
xlabel style= {font=\color{white!15!black}},
x label style={at={(axis description cs:0.5, -0.02)},anchor=north},
xlabel={$x_1$},
ymin=30,
ymax=95,
ylabel style={font=\color{white!15!black}},
y label style={at={(axis description cs:-0.02, 0.5)},anchor=south},
ylabel={$x_2$},
axis background/.style={fill=white},
label style={font=\Large},
tick style={draw=none},
yticklabel=\empty,
xticklabel=\empty,
]

\addplot[area legend, line width=0pt, fill=darkgray, fill opacity=0.25]
table[row sep=crcr] {%
x	y\\
40	40\\
40	60\\
60	60\\
60	40\\
}--cycle;
\addlegendentry{actual rectangle1}


\addplot[area legend, line width=3.5pt, dashed, draw=red]
table[row sep=crcr] {%
x	y\\
39.0791392703541	40\\
39.0791392703541	59.9999999999986\\
60	59.9999999999986\\
60	40\\
}--cycle;
\addlegendentry{predicted rectangle1}

\addlegendentry{Predicted points}

\addplot [color=gray2, mark=*, mark options={solid, gray2, fill}]
  table[row sep=crcr]{%
10	50\\
10.0735362623468	49.9707867663623\\
10.2910649504412	49.8848552995644\\
10.6464280045546	49.7456166087229\\
11.133471545309	49.5564711120904\\
11.7460456841466	49.3208092888785\\
12.4780043323592	49.0420123314026\\
13.3232050086939	48.7234527946081\\
14.2755086488295	48.368495244439\\
15.3287794114454	47.9804969049938\\
16.4768844871305	47.5628083055199\\
17.7136939038841	47.118773924093\\
19.0330803360245	46.6517328328337\\
20.428918906851	46.165019341101\\
21.895086997836	45.6619636372327\\
23.425464052707	45.1458924319821\\
25.0139313829472	44.6201295981559\\
26.6543719741493	44.0879968130806\\
28.3406702881088	43.5528141981599\\
30.066712071449	43.0179009598053\\
31.826384156295	42.4865760290737\\
33.6135742674786	41.9621587025176\\
35.4221708254843	41.4479692818228\\
37.2460627516664	40.9473297143349\\
39.0791392703427	40.4635642335835\\
40.9152897151482	40\\
42.7484033322643	39.5590297169522\\
44.572369081945	39.1411740669456\\
46.3810754468574	38.7460198404319\\
48.1684102333242	38.373157885351\\
49.9282603751209	38.0221830317017\\
51.6545117386935	37.6926940166615\\
53.3410489278073	37.3842934103525\\
54.9817550859255	37.0965875420704\\
56.5705117028456	36.8291864271719\\
58.1011984173617	36.5817036945423\\
59.5676928241851	36.3537565144775\\
60.9638702766061	36.1449655271574\\
62.2836036948351	35.9549547717304\\
63.520763367812	35.7833516157219\\
64.6692167615473	35.6297866851289\\
65.7228283251908	35.4938937948191\\
66.6754592956816	35.3753098795854\\
67.5209675066559	35.2736749254897\\
68.2532071933809	35.1886319018528\\
68.8660288027988	35.1198266934969\\
69.3532787984601	35.0669080335592\\
69.7087994717039	35.0295274366696\\
69.9264287485904	35.0073391325332\\
70	35\\
};
\addlegendentry{optimal path 1}

\addplot [color=blue, only marks, mark=o, mark options={solid, blue}, forget plot]
  table[row sep=crcr]{%
10	50\\
};
\addplot [color=red, only marks, mark=o, mark options={solid, red}, forget plot]
  table[row sep=crcr]{%
70	35\\
};
\addplot [color=gray2, mark=*, mark options={solid, gray2, fill}]
  table[row sep=crcr]{%
10	90\\
10.0814078046405	89.9840577842985\\
10.3219843336208	89.9353440600632\\
10.7144381791331	89.8520811100695\\
11.251482544694	89.7324839207108\\
11.9258350209992	89.5747598404543\\
12.7302173581883	89.3771082373051\\
13.6573552395353	89.1377201547104\\
14.6999780568521	88.8547779660481\\
15.8508186807195	88.5264550290433\\
17.1026132333592	88.1509153366036\\
18.448100859492	87.7263131680795\\
19.8800235014758	87.2507927381033\\
21.391125665069	86.7224878439528\\
22.9741541929997	86.1395215119652\\
24.6218580366453	85.500005641168\\
26.3269880227177	84.8020406459707\\
28.0822966261994	84.0437150968986\\
29.8805377383648	83.2231053595398\\
31.7144664378097	82.3382752314097\\
33.5768387570591	81.3872755771121\\
35.460411455668	80.3681439613383\\
37.3579417861461	79.2789042798379\\
39.2621872637126	78.1175663883536\\
41.1659054355923	76.882125729349\\
43.0618536508097	75.5705629566345\\
44.9427888271535	74.1808435572441\\
46.8014672218434	72.7109174717361\\
48.6306441995239	71.1587187114977\\
50.4230740017303	69.5221649737432\\
52.1715095153124	67.7991572540729\\
53.8687020446537	65.987579456689\\
55.5074010762704	64.0852980013329\\
57.0803540531637	62.0901614282719\\
58.5803061407783	60\\
60	57.8211619079077\\
61.3336055718151	55.5770511118074\\
62.5781499078673	53.2995952357952\\
63.7320870570498	51.0207150952839\\
64.7938679971181	48.7723308337283\\
65.7619405813975	46.5863680544952\\
66.6347494859975	44.4947639624451\\
67.4107361584192	42.5294734964458\\
68.0883387646712	40.722475470587\\
68.6659921384318	39.1057787229897\\
69.1421277296372	37.7114282564967\\
69.5151735535857	36.5715113979155\\
69.7835541404951	35.7181639571357\\
69.9456904858396	35.1835763951222\\
70	35\\
};
\addlegendentry{optimal path 2}

\addplot [color=blue, only marks, mark=o, mark options={solid, blue}, forget plot]
  table[row sep=crcr]{%
10	90\\
};
\addplot [color=red, only marks, mark=o, mark options={solid, red}, forget plot]
  table[row sep=crcr]{%
70	35\\
};
\legend{}
\end{axis}
\end{tikzpicture}%
        \caption{$\nb = 1, L = 2$}
        \label{fig:case1}
    \end{subfigure}
    \begin{subfigure}[t]{0.24\textwidth}
        \input{images/QR_analysis/case_2_pred_case}
        \caption{$\nb = 2, L = 4$}
        \label{fig:case2}
    \end{subfigure}
    \begin{subfigure}[t]{0.24\textwidth}
        \input{images/QR_analysis/case_3_pred_case}
        \caption{$\nb = 2, L = 4$}
        \label{fig:case3}
    \end{subfigure}
    \begin{subfigure}[t]{0.24\textwidth}
%
\definecolor{mycolor1}{rgb}{0.90000,0.27000,0.27000}%
\definecolor{mycolor2}{rgb}{0.27000,0.90000,0.27000}%
\definecolor{mycolor3}{rgb}{0.27000,0.27000,0.90000}%

\definecolor{gray2}{rgb}{0.4,0.4,0.4}%
\definecolor{black2}{rgb}{0.0,0.0,0.0}%

\begin{tikzpicture}[scale=0.35]

\begin{axis}[%
width=4.5in,
height=4.5in,
at={(0.751in,1.253in)},
scale only axis,
xmin=-7.02120099638328,
xmax=102.0212009963833,
xlabel style= {font=\color{white!15!black}},
x label style={at={(axis description cs:0.5, -0.02)},anchor=north},
xlabel={$x_1$},
ymin=0,
ymax=90.115700926553,
ylabel style={font=\color{white!15!black}},
y label style={at={(axis description cs:-0.02, 0.5)},anchor=south},
ylabel={$x_2$},
label style={font=\Large},
axis background/.style={fill=white},
tick style={draw=none},
yticklabel=\empty,
xticklabel=\empty,
]

\addplot[area legend, draw=black, fill=darkgray, fill opacity=0.25]
table[row sep=crcr] {%
x	y\\
10	10\\
10	30\\
30	30\\
30	10\\
}--cycle;
\addlegendentry{actual rectangle1}

\addplot[area legend, draw=black, fill=darkgray, fill opacity=0.25]
table[row sep=crcr] {%
x	y\\
45	45\\
45	75\\
75	75\\
75	45\\
}--cycle;
\addlegendentry{actual rectangle2}

\addplot[area legend, draw=black, fill=darkgray, fill opacity=0.25]
table[row sep=crcr] {%
x	y\\
31.5	64.5\\
31.5	71.5\\
38.5	71.5\\
38.5	64.5\\
}--cycle;
\addlegendentry{actual rectangle3}

\addplot[area legend, line width=3.5pt, dashed, draw=red]
table[row sep=crcr] {%
x	y\\
45	45\\
45	75\\
75.0000000000051	75\\
75.0000000000051	45\\
}--cycle;
\addlegendentry{predicted rectangle1}

\addplot[area legend, line width=3.5pt, dashed, draw=red]
table[row sep=crcr] {%
x	y\\
10	10.0000000000027\\
10	30.0000000001189\\
29.9999999999896	30.0000000001189\\
29.9999999999896	10.0000000000027\\
}--cycle;
\addlegendentry{predicted rectangle2}

\addplot[area legend, line width=3.5pt, dashed, draw=red]
table[row sep=crcr] {%
x	y\\
31.5	64.5\\
31.5	71.5\\
38.5	71.5\\
38.5	64.5\\
}--cycle;
\addlegendentry{predicted rectangle3}

\addplot [color=gray2, mark=*, mark options={solid, gray2, fill}]
  table[row sep=crcr]{%
5	5\\
5.08422774248051	5.00057941389908\\
5.33337725391122	5.00458727216747\\
5.74038512394979	5.01656293775886\\
6.29819580830763	5.04104614738558\\
6.9997612739464	5.08257723763751\\
7.83804064546712	5.14569737240578\\
8.80599984485437	5.23494876959387\\
9.89661123873204	5.35487492790721\\
11.1028532842174	5.51002085357618\\
12.4177101720124	5.70493328629334\\
13.8341714638741	5.94416092860047\\
15.3452317430638	6.23225466809584\\
16.9438902509864	6.57376780643325\\
18.6231505328324	6.97325628553517\\
20.3760200799443	7.43527891254053\\
22.195509969136	7.96439758653287\\
24.0746345078664	8.56517752457876\\
26.0064108737527	9.24218748608288\\
27.9838587565642	10\\
30	10.8413068419322\\
32.0483497623982	11.7650344198739\\
34.1234077067156	12.7682284620823\\
36.2201661907602	13.847938611405\\
38.3336184876971	15.0012182721957\\
40.4587585240656	16.2251244582238\\
42.5905806198716	17.5167176394025\\
44.7240792254452	18.8730615888724\\
46.85424866213	20.291223229724\\
48.9760828623548	21.7682724812767\\
51.0845751066227	23.3012821053853\\
53.1747177639045	24.8873275540536\\
55.2415020302869	26.5234868135717\\
57.279917667997	28.2068402515995\\
59.284952743561	29.9344704630516\\
61.2515933691612	31.7034621154806\\
63.1748234442202	33.5109017948853\\
65.0496243842645	35.3538778509246\\
66.8709748730599	37.2294802425895\\
68.6338505941641	39.1348003831226\\
70.3332239743696	41.066930986126\\
71.9640639218269	43.0229659090831\\
73.5213355664341	45\\
75	46.9939073606508\\
76.3967777091653	48.9981195495379\\
77.7119120969549	51.0048468244929\\
78.9474057952626	53.006299325265\\
80.1052570654339	54.9946866692044\\
81.1874598918366	56.9622175593395\\
82.1960040765698	58.9010993796401\\
83.1328753325506	60.8035377960186\\
84.0000553778955	62.6617363591213\\
84.7995220311939	64.467896102462\\
85.5332493050096	66.2142151448632\\
86.2032075010939	67.8928882907164\\
86.8113633052709	69.4961066313994\\
87.3596798823099	71.0160571460146\\
87.8501169708312	72.4449223028496\\
88.2846309789884	73.7748796611\\
88.6651750796009	74.9981014729708\\
88.9936993054595	76.1067542839107\\
89.2721506451904	77.0929985360708\\
89.50247313887	77.9489881720794\\
89.6866079738288	78.6668702358889\\
89.8264935803018	79.2387844765008\\
89.9240657283495	79.6568629519218\\
89.9812576228476	79.9132296312721\\
90	80\\
};
\addlegendentry{optimal path 1}

\addplot [color=gray2, mark=*, mark options={solid, gray2, fill}]
  table[row sep=crcr]{%
5	20\\
5.02926926702163	20.0627267814093\\
5.11715550775009	20.2480979537198\\
5.26381705621486	20.5504981186184\\
5.46941515954159	20.9643176278741\\
5.73411398530934	21.4839523013858\\
6.05808062871631	22.1038031448368\\
6.44148511958276	22.8182760670904\\
6.88450042926592	23.6217815974767\\
7.38730247738949	24.5087346027535\\
7.95007013841198	25.4735540037509\\
8.57298524806379	26.5106624921203\\
9.25623260960923	27.6144862466168\\
10	28.7794546489996\\
10.8036069909888	30.0000000001115\\
11.6646337265847	31.2712372262372\\
12.5797919827696	32.5896436195386\\
13.5457961838211	33.9523786648376\\
14.5593632332748	35.3566039006355\\
15.6172123450087	36.7994827691\\
16.716064874432	38.2781804692931\\
17.852644148373	39.7898638069164\\
19.023675296572	41.3317010454406\\
20.225885083084	42.9008617573064\\
21.4560017341283	44.4945166737307\\
22.7107547710301	46.1098375367888\\
23.9868748394848	47.7439969491403\\
25.2810935404903	49.3941682252334\\
26.5901432583084	51.0575252403296\\
27.9107569940388	52.7312426629639\\
29.2396681934467	54.4124942832452\\
30.5736105778487	56.0984547648113\\
31.9093179733027	57.7862994968148\\
33.24352414153	59.4732025390456\\
34.5729626098351	61.1563379967119\\
35.8943665012652	62.8328794890755\\
37.204468363575	64.5\\
38.5	66.1519786408455\\
39.7818726982142	67.7773072696287\\
41.0593576674642	69.3615829942775\\
42.3459060969641	70.8904005293734\\
43.6549694235712	72.3493514726872\\
45	73.7240230905938\\
46.3899210787299	75\\
47.8245965765748	76.1675763113538\\
49.2993617370341	77.2264776330676\\
50.8095535699979	78.1811409619747\\
52.3505106096855	79.0359976970106\\
53.917572674181	79.7954738559963\\
55.5060806254623	80.4639902926645\\
57.1113761279863	81.0459629138975\\
58.7288014082364	81.5458028965779\\
60.3536990134948	81.9679169076245\\
61.9814115710069	82.3167073182987\\
63.6072815471302	82.5965724241679\\
65.2266510059148	82.8119066634898\\
66.8348613683553	82.9671008352124\\
68.4272531716852	83.0665423167777\\
69.9991658281082	83.1146152834528\\
71.5459373829653	83.115700926553\\
73.0629042754464	83.0741776730273\\
74.545401096515	82.9944214038551\\
75.988760348293	82.8808056725165\\
77.3883122035768	82.7377019246293\\
78.7393842646077	82.5694797177566\\
80.0373013229195	82.3805069390793\\
81.2773851186355	82.1751500257444\\
82.4549541013917	81.9577741839455\\
83.5653231860811	81.7327436079341\\
84.6038035150886	81.5044216993774\\
85.565702222288	81.2771712867839\\
86.4463221880402	81.0553548447139\\
87.2409618002467	80.8433347125471\\
87.9449147158576	80.6454733156486\\
88.5534696216035	80.4661333816272\\
89.0619099944366	80.3096781612296\\
89.465513863387	80.1804716474203\\
89.7595535702742	80.0828787940172\\
89.939295531519	80.0212657342377\\
90	80\\
};
\addlegendentry{optimal path 2}

\addplot [color=gray2, mark=*, mark options={solid, gray2, fill}]
  table[row sep=crcr]{%
20	50\\
20.0192664789797	50.1000357809677\\
20.0792152724258	50.3926331878004\\
20.1841461141382	50.8627768375898\\
20.3383608865551	51.4954599811379\\
20.5461638353055	52.275683738843\\
20.8118617832661	53.1884563504718\\
21.1397643460439	54.2187924216915\\
21.5341841455135	55.3517121706393\\
21.999437022605	56.5722406729875\\
22.5398422527997	57.8654072211376\\
23.1597227584571	59.2162440009415\\
23.8634053233744	60.6097864782166\\
24.655220806345	62.031071500387\\
25.5395043517929	63.4651371119544\\
26.5205956053141	64.8970216880552\\
27.6028389245601	66.3117631765969\\
28.7905831039435	67.6943983449597\\
30.0881840279588	69.029962021553\\
31.5000000000001	70.303486342111\\
33.0270059011391	71.5\\
34.6633998546749	72.6091354103267\\
36.3999943519011	73.6297361175811\\
38.2276067207567	74.5652488038177\\
40.1370586582925	75.4191155289299\\
42.1191757612396	76.1947738978779\\
44.1647870604827	76.8956572280731\\
46.2647245503031	77.525194717021\\
48.4098227198701	78.0868116104736\\
50.5909181778447	78.5839293710312\\
52.7988487210729	79.0199658459751\\
55.0244537882973	79.3983354356057\\
57.2585730713326	79.7224492618116\\
59.492046502186	79.9957153366444\\
61.7157136954897	80.2215387313225\\
63.9204134775792	80.4033217448746\\
66.0969834184239	80.5444640736687\\
68.2362593617291	80.6483629796617\\
70.3290749562426	80.7184134603792\\
72.3662611860389	80.7580084165784\\
74.3386459022338	80.7705388241448\\
76.2370533546075	80.7593939015458\\
78.0523037225954	80.7279612795683\\
79.7752126450663	80.6796271706873\\
81.396590756312	80.6177765387116\\
82.9072432142652	80.5457932674515\\
84.2979692316144	80.4670603324119\\
85.5595616130571	80.3849599694251\\
86.6828062843527	80.3028738418215\\
87.6584818262516	80.2241832120883\\
88.4773590078174	80.1522691114935\\
89.1302003205134	80.0905125089537\\
89.6077595110028	80.042294480361\\
89.9007811164099	80.0109963780047\\
90	80\\
};
\addlegendentry{optimal path 3}

\legend{}
\end{axis}
\end{tikzpicture}%
        \caption{$\nb = 3, L = 3$}
        \label{fig:case4}
    \end{subfigure}\\[1em]

    \begin{subfigure}[t]{0.24\textwidth}
        \input{images/QR_analysis/case_5_pred_case}
        \caption{$\nb = 2, L = 5$} 
        \label{fig:case5}
    \end{subfigure}
    \begin{subfigure}[t]{0.24\textwidth}
        \input{images/QR_analysis/case_6_pred_case}
        \caption{$\nb = 2, L = 6$}
        \label{fig:case6}
    \end{subfigure}
    \begin{subfigure}[t]{0.24\textwidth}
        \input{images/QR_analysis/case_7_pred_case}
        \caption{$\nb = 3, L = 4$}
        \label{fig:case7}
    \end{subfigure}
    \begin{subfigure}[t]{0.24\textwidth}
        \input{images/QR_analysis/case_8_pred_case}
        \caption{$\nb = 2, L = 4$}
        \label{fig:case8}
    \end{subfigure}
    \caption{The environments used in the simulation experiments. Each
    subcaption indicates the number of unknown constraints ($\nb$) and
    demonstrations ($L$) for that scenario. Legend: {\protect\drawgt} obstacles,
    {\protect\drawtraj} optimal trajectory, and {\protect\drawpred} learned
    constraints.}
    \label{fig:QR_exp}
    \figvspace{}
\end{figure*}

The purpose of these experiments is to:
\begin{itemize}
    \item study the importance of learning the cost weight in the context of
    constraint learning with an unknown policy;
    \item evaluate the ability of the proposed method to jointly learn both
    unknown cost and constraints with varying number of demonstrations and
    unknown constraints;
    \item demonstrate how the proposed method can be used on real trajectories
    in a robotic manipulation setting.
\end{itemize}

First, we will present a series of quantitative experiments from a set of
eight simulated environments with varying
number of demonstrations and unknown constraints. In simulation, we will
present our full method together with an evaluation of the proposed method to
extract constraints when faced with user-defined cost with different degrees of
error in the cost parametrization. This aims to illustrate the implications of
user-selected cost settings on constraint learning processes and the impact of
jointly learning cost and constraints. Finally, a real manipulation task with a
robotic arm is presented.    
\subsection{Simulation experiment setting}
\label{sec:sim_settings}
For our simulation setting, we consider an obstacle avoidance problem for a
four-dimensional free-floating system, assuming unknown cost and obstacles but
with known system dynamics, as well as velocity and acceleration constraints.

\begin{figure*}
    \centering
    \begin{subfigure}[t]{0.32\textwidth}
        \input{images/real_world_exp/outliers}
        \caption{Demonstrations with detected outliers and learned constraint.}
        \label{fig:man_out}
    \end{subfigure}
    \hfill
    \begin{subfigure}[t]{0.32\textwidth}
%
\definecolor{mycolor1}{rgb}{0.90000,0.27000,0.27000}%
\definecolor{mycolor2}{rgb}{0.27000,0.90000,0.27000}%
\definecolor{mycolor3}{rgb}{0.27000,0.27000,0.90000}%
\definecolor{gray2}{rgb}{0.4,0.4,0.4}%
\definecolor{black2}{rgb}{0.0,0.0,0.0}%
\begin{tikzpicture}[scale = 0.62]

\begin{axis}[%
width=3.0in,
height=2.5in,
at={(0.525in,0.486in)},
scale only axis,
xmin=-0.2,
xmax=0.4,
xlabel style={font=\color{white!15!black}},
xlabel={$y$},
ymin=2.8,
ymax=4.8,
ylabel style={font=\color{white!15!black}},
ylabel={$\theta$},
axis background/.style={fill=white},
label style={font=\Large},
axis x line*=bottom,
axis y line*=left,
legend style={at={(1.03,1)}, anchor=north west, legend cell align=left, align=left, draw=white!15!black}
]

\addplot[area legend, line width=2.5pt, dashed, draw=red, pattern=north west lines, pattern color=red]
table[row sep=crcr] {%
x	y\\
-9.70221251947686	-5.38085896789839\\
-9.70221251947686	4.637764446376\\
0.258544695529116	4.637764446376\\
0.258544695529116	-5.38085896789839\\
}--cycle;

\addlegendentry{actual rectangle1}

\addplot [color=green, line width=2.0pt]
  table[row sep=crcr]{%
-0.0511	4.645\\
-0.0376473725828413	4.64533799969781\\
-0.0107691748792408	4.64601399909343\\
0.0160849364338467	4.64668999858448\\
0.0429750238175511	4.6473659980435\\
0.0699612304801805	4.64804199750252\\
0.0971039146134218	4.64871799696154\\
0.124463784391354	4.64939399642056\\
0.152102033752246	4.65006999587957\\
0.180080479266823	4.65074599533859\\
0.208461698399143	4.65142199361923\\
0.237309169314104	4.63952999999999\\
0.258499999999998	4.60250202494669\\
0.263894173519918	4.55290605916472\\
0.261691167660526	4.503310089831\\
0.260073468275279	4.45371412169169\\
0.259037457322463	4.40411815355237\\
0.258580817623823	4.35452218541305\\
0.258499999999998	4.30492621727374\\
0.258499999999998	4.25533024913442\\
0.258499999999998	4.20573428101676\\
0.258499999999998	4.15613831287745\\
0.258499999999998	4.10654234473813\\
0.258499999999998	4.05694637659881\\
0.258499999999998	4.00735040847931\\
0.258499999999998	3.95775444033999\\
0.258499999999998	3.90815847220068\\
0.258499999999998	3.85856250406136\\
0.258499999999998	3.80896653592204\\
0.258499999999998	3.75937056778273\\
0.258499999999998	3.70977459964341\\
0.258499999999998	3.6601786315041\\
0.258499999999998	3.61058266336478\\
0.258499999999984	3.56098669522547\\
0.258610702021487	3.51139072813517\\
0.259127177356425	3.46179475999585\\
0.260223224616732	3.41219879185653\\
0.261901295259176	3.36260282371722\\
0.264165142508905	3.3130068555779\\
0.267019829757301	3.26341088743859\\
0.270471741886918	3.21381491929927\\
0.274528599552113	3.16421895115996\\
0.279199476447298	3.11462298302064\\
0.284494819523609	3.0762125\\
0.2873	3.0626\\
};
\addlegendentry{optimal path 1}

\addplot [color=green, line width=2.0pt]
  table[row sep=crcr]{%
-0.01	4.685\\
0.00361250000000002	4.6829955312906\\
0.0314631722776983	4.67898659368208\\
0.0599906253473373	4.67497765588731\\
0.0886522553563184	4.67096871809254\\
0.117512167765238	4.66695978029778\\
0.146634911554505	4.66295084250301\\
0.176085623561399	4.65894190470824\\
0.205930174167143	4.65493296584236\\
0.236235314462219	4.63952999999999\\
0.258499999999998	4.60133898233795\\
0.264205202964319	4.55175393876423\\
0.261932509496514	4.5021688919848\\
0.260245662288302	4.45258384627394\\
0.259140888647024	4.40299880056308\\
0.258615717598284	4.35341375485222\\
0.258499999999998	4.30382870914136\\
0.258499999999998	4.2542436634305\\
0.258499999999998	4.20465861778661\\
0.258499999999998	4.15507357207575\\
0.258499999999998	4.10548852636489\\
0.258499999999998	4.05590348065403\\
0.258499999999998	4.00631843496658\\
0.258499999999998	3.95673338925572\\
0.258499999999998	3.90714834354486\\
0.258499999999998	3.85756329783401\\
0.258499999999998	3.80797825212315\\
0.258499999999998	3.75839320641229\\
0.258499999999998	3.70880816071045\\
0.258677211978791	3.65922311499959\\
0.259316724418434	3.60963806928873\\
0.260536232731387	3.56005302357787\\
0.26233846451019	3.51046797786701\\
0.264727450682664	3.46088293215615\\
0.267708534529807	3.4112978874938\\
0.271288383636751	3.36171284178294\\
0.275475004805687	3.31212779607208\\
0.280277761964126	3.26254275036122\\
0.285707397108557	3.21295770465036\\
0.291776054330328	3.1633726589395\\
0.298497306977499	3.11378761322864\\
0.305886188013418	3.06420256751778\\
0.313959223639915	3.01461752180693\\
0.322734470138912	2.9762125\\
0.3273	2.9626\\
};
\addlegendentry{optimal path 2}

\addlegendentry{optimal path 3}

\legend{};
\end{axis}
\end{tikzpicture}%
        \caption{Generated trajectories from the learned cost and constraint.}
        \label{fig:man_gen_traj}
    \end{subfigure}
    \hfill
    \begin{subfigure}[t]{0.32\textwidth}
        \definecolor{mycolor1}{rgb}{0.90000,0.27000,0.27000}%
\begin{tikzpicture}[scale=0.62]

\begin{axis}[%
width=3.0in,
height=2.5in,
at={(0.78in,0.478in)},
scale only axis,
xmin=0,
xmax=30,
ymin=-0.0005,
ymax=5.9e+09,
axis background/.style={fill=white},
ylabel style={font=\color{white!15!black}},
ylabel={$\text{Tr}(\NQ^j)$},
xlabel style={font=\color{white!15!black}},
xlabel={$j$},
label style={font=\Large},
axis x line*=bottom,
axis y line*=left,
legend style={legend cell align=left, align=left, draw=white!15!black}
]

\addplot [color=brown, line width=2.0pt]
  table[row sep=crcr]{%
1    0.0000e+09 \\    
2    0.0000e+09 \\    
3    0.0000e+09 \\    
4    0.0000e+09 \\    
5    5.8746e+09 \\    
6    0.0000e+09 \\    
7    0.0000e+09 \\    
8    0.0000e+09 \\    
9    0.0000e+09 \\    
10    0.0000e+09 \\    
11    0.0000e+09 \\    
12    0.0000e+09 \\    
13    0.0847e+09 \\    
14    0.0000e+09 \\    
15    0.0000e+09 \\    
16    0.0000e+09 \\    
17    0.0000e+09 \\    
18    0.0124e+09 \\    
19    0.0000e+09 \\    
20    0.0000e+09 \\    
21    0.0000e+09 \\    
22    0.0000e+09 \\    
23    0.0000e+09 \\    
24    0.0000e+09 \\    
25    0.0000e+09 \\    
26    0.0000e+09 \\    
27    0.5684e+09 \\    
28    0.0000e+09 \\
};
\addlegendentry{$\text{obj}_\text{v}\text{al}$}

\addplot [color=red, only marks, mark=o,  line width=1.0pt, mark size=4pt, mark options={solid, red}]
  table[row sep=crcr]{%
5 5.8746e+09\\
};

\addplot [color=green, only marks, mark=o, line width=1.0pt, mark size=4pt, mark options={solid, green}]
  table[row sep=crcr]{%
13	0.0847e+09\\
};

\addplot [color=red, only marks, mark=o, line width=1.0pt, mark size=4pt, mark options={solid, red}]
  table[row sep=crcr]{%
27	0.5684e+09\\
};
\addlegendentry{outliers 4}
\legend{}
\end{axis}
\end{tikzpicture}%
        \caption{$\text{Tr}(\NQ^j)$ for one demonstration, showing a failure case of outlier detection.} 
        \label{fig:man_obj_out}
    \end{subfigure}
    \caption{Real manipulator experiment. Legend: {\protect\drawtraj}
    collected demonstration, {\protect\drawout} outlier, {\protect\drawpredman}
    learned constraint, {\protect\drawgentraj} generated trajectory, {\protect\drawtraceQ}
    trace of normalized $Q^j$, {\protect\drawQoutFP} false positive,  {\protect\drawQoutTP} true positive.}
    \label{fig:man_exp}
    \figvspace{}
\end{figure*}

The initial and final position constraint for each demonstration $\Zl$:
\begin{equation}    
\begin{aligned}
\xe{}{l}{1} - \xs^l = 0, \quad  
\xe{}{l}{N_l} - \xg^l = 0 \enspace.
\end{aligned}
\end{equation}
System dynamics constraint can be represented as:
\begin{equation}
\begin{gathered}
\xl{k+1}  = A\xl{k}  + B\ull{k}  \\
A = 
\begin{bmatrix}
    1 & 0 & \mathrm{dt} & 0\\ 
    0 & 1 & 0 & \mathrm{dt}\\ 
    0 & 0 & 1 & 0\\ 
    0 & 0 & 0 & 1
\end{bmatrix}, \quad
B = 
\begin{bmatrix}    
0.5\,{\mathrm{dt}}^2 & 0\\ 
0 & 0.5\,{\mathrm{dt}}^2\\
\mathrm{dt} & 0\\
0 & \mathrm{dt}
\end{bmatrix},
\end{gathered}
\label{eq:free_floating}
\end{equation}
where, $\xl{} = [\xe{1}{l}{}, \xe{2}{l}{}, \xe{3}{l}{}, \xe{4}{l}{}]^{\top}$
represents the state vector, while $\ull{} = [\ue{1}{l}{}, \ue{2}{l}{}]^{\top}$
denotes the control vector. Specifically, $\xe{1}{l}{}$ and $\xe{2}{l}{}$
correspond to positions along the $x$-axis and $y$-axis, respectively, whereas
$\xe{3}{l}{}$ and $\xe{4}{l}{}$ represent velocities in the $x$-axis and
$y$-axis. Control inputs $\ue{1}{l}{}$ and $\ue{2}{l}{}$ are accelerations in
the $x$-axis and $y$-axis directions, respectively.

In each of the eight simulation scenarios, shown in \figref{fig:QR_exp}, we
introduce a number of inequality constraints representing obstacles. Specifically, in each scenario we
include a different number $\nb$ of unknown axis-aligned exclusive constraints
for the positional states $\xe{1}{l}{}$ and $\xe{2}{l}{}$, aimed at avoiding
obstacles, as well as known inclusive constraints for the velocity states and acceleration input
$\xe{3}{l}{}$ and $\xe{4}{l}{}$, which ensure the velocities and accelerations remain within
predefined bounds. In each scenario we generate $L$ optimal demonstrations according to a single $Q$ and $R$ per scenario and the system dynamics presented above; each demonstration has a different starting and goal point.

For our approach, in the context of cost learning, we define the vectors for states as $\vpx = [0, 0, 0, 0]^{\top}$ and $\vsx = [0, 0, 1, 1]^{\top}$, and for controls as $\vpu = [0, 0]^{\top}$ and $\vsu = [1, 1]^{\top}$. In the case of learning constraints, we set for states $\vpx = [1, 1, 0, 0]^{\top}$ and $\vsx = [1, 1, 1, 1]^{\top}$, and for controls $\vpu = [0, 0]^{\top}$ and $\vsu = [1, 1]^{\top}$.

\subsection{Influence of cost weights on the retrieved constraints}\label{sec:cost_weight_influence}

To investigate how cost impacts the constraints extractions, we pair the method
used for extracting constraints with eight different user-defined $Q$, where in
each alter two components of the weight matrix $Q$. This modification is
intended to mimic the real-world scenario of manually adjusting the weights,
illustrating the practical challenges encountered in the demonstrations. 

Initially, for each environment tested, we create distinct sets of cost $Q$ and
$R$, ensuring that $Q \geq 0$, $\text{Tr}(Q) = 1$, and $R$ is strictly positive definite.
Subsequently, $L$ demonstrations are conducted using these predetermined $Q$ and
$R$ values. To mimic the real-world practice of guessing weight for the
demonstrations, we introduce perturbed versions of $Q$, denoted as
$\tilde{Q}_i$, generated according to: 
\begin{equation*}
\begin{aligned}
    &\tilde{Q}_i = diag([Q(1,1) + \sigma_i, Q(2,2) - \sigma_i, Q(3,3), Q(4,4)]), \\
    &\sigma_i \in \{-0.01, -0.005, -0.001, 0.0, 0.001, 0.005, 0.01, 0.02\}.
\end{aligned}
\end{equation*}
We then perform constraint extraction according to \secref{sec:cons_ext}.
Moreover, to evaluate our proposed method, we run the full proposed method,
performing also cost extraction according to \secref{sec:cost_ext} together with
constraints extraction.

\begin{table}
\centering
\caption{Average results over eight simulated scenarios}
\label{tab:QR_analysis}
\begin{tabular}{HcHcccccc}
\toprule
\textbf{cases} & \textbf{$\sigma$} & \textbf{N} & \textbf{Conv. $\%$} & \textbf{RMSE} & \textbf{Prec.} & \textbf{Recall} & \textbf{F1} & \textbf{IoU} \\
\midrule
12364895.0000 & -0.010 & 8.0000 & 37.5 & 0.024 & 0.667 & 0.494 & 0.994 & 0.494 \\
12364895.0000 & -0.005 & 8.0000 & 37.5 & 0.012 & 0.993 & 0.750 & 0.996 & 0.744 \\
12364895.0000 & -0.001 & 8.0000 & 50.0 & 0.002 & 0.750 & 0.744 & 0.996 & 0.744 \\
\rowcolor[gray]{0.93} 
12364895.0000 & \phantom{-}0.000 & 8.0000 & 100.0 & 0.000 & 0.994 & 1.000 & 0.997 & 0.994 \\
12364895.0000 & \phantom{-}0.001 & 8.0000 & 75.0 & 0.002 & 0.869 & 0.875 & 0.997 & 0.869 \\
12364895.0000 & \phantom{-}0.005 & 8.0000 & 37.5 & 0.012 & 0.565 & 0.500 & 0.994 & 0.494 \\
12364895.0000 & \phantom{-}0.010 & 8.0000 & 37.5 & 0.024 & 0.391 & 0.250 & 0.987 & 0.244 \\
12345689.0000 & \phantom{-}0.020 & 8.0000 & 12.5 & 0.047 & 0.391 & 0.247 & 0.983 & 0.242 \\
\midrule
12364895 & Learned & 8 & 100.0 & 7.13e-13 & 0.994 & 1.000 & 0.997 & 0.994 \\
\bottomrule
\end{tabular}
\figvspace{}
\end{table}

\tabref{tab:QR_analysis} presents the average results of these tests over the
eight simulation scenario. We categorize the solution to the problem
\eqref{eq:cons_extraction} as converged (Conv.) if it identifies a feasible
solution within 10 minutes. The
Root Mean Square Error (RMSE$(y, y^*)$) is used to quantify the deviation
between known and perturbed costs. The vector $y(Q, R)$ is defined as
$\left[\frac{Q(1,1)}{R(1,1)}, \frac{Q(2,2)}{R(2,2)}, \frac{Q(3,3)}{R(1,1)},
\frac{Q(4,4)}{R(2,2)}\right]^{\top}$, and similarly $y^*(\Qs, \Rs)$ is
formulated for the perturbed weights. This specific configuration is selected
because the control input along the $x$-axis influences the position and
velocity in the $x$-axis, and likewise, the control input along the $y$-axis
affects the position and velocity in the $y$-axis. All other metrics refer to
the area of the retrieved constraints when compared with the ground-truth
constraints shown in \figref{fig:QR_exp}. The performance of our constraint
learning method on ground-truth cost is indicated in light gray, while row
indicated as ``Learned'' denotes the results when both our constraints learning
and our cost learning methods are used\add{\footnote{It is worth mentioning that
some trajectories intersect the borders. This occurs due to the sampling time used;
smaller sampling times generate trajectories closer to the boundary. In practice,
constraints can be expanded to enhance safety.}}.

These results reveal the significant influence of perturbed cost
parameter $\Qs$ on both convergence rates and overall solution quality
within a constraint learning framework. Notably, when deviations in $\Qs$ are
confined to $\pm 0.005$, a mere $37.5\%$ of the scenarios reach convergence.
Additionally, among these scenarios, a marked reduction in recall to $0.75$ is
observed. Despite this, consistently high F1 scores in a majority of cases
indicate the efficacy of the methodology in maintaining a balance between
precision and recall. However, precision, recall, and Intersection over Union
(IoU) metrics show a decline as the deviations in $\Qs$ widen. This pattern
highlights the sensitivity of the constraint learning process to slight
alterations in specific $\Qs$ parameters. Therefore, accurately learning the
unknown cost is a critical step in our proposed method. It not only facilitates
the acquisition of skills from demonstrations but also significantly influences
the constraints affecting those skills, highlighting that accurate estimation of
cost parameters is pivotal for learning skills under constraints. The last row in \tabref{tab:QR_analysis}
shows that jointly learning cost and constraints closely mirrors the known
cost and constraints performance.

\subsection{Robotic manipulation experiment}

In this experiment, a user performs a set of demonstrations hand-guiding a Franka
Emika Panda 7-DoF robotic manipulator holding a cup in the end effector. The
task is to drop a ball contained in the cup into a bucket. This setup is
depicted in \figref{fig:cover}. The trajectory followed during these
demonstrations is illustrated in \figref{fig:man_out}.

In these demonstrations, $x_1$ corresponds to the displacement along the y-axis,
and $x_2$ represents the angle of rotation, within the global frame.
Additionally, $x_3$ and $x_4$ are associated with positional and angular
velocity, respectively. All other settings follow the free-floating system
presented in \secref{sec:sim_settings}.

In \figref{fig:man_out}, a high number of detected outlier can be noted. Many of the outliers detected are
false positive due to the real demonstrations being locally suboptimal in those
spots. An example demonstration where this phenomenon is visible is shown in \figref{fig:man_obj_out}, where we can note how the trace of the normalized cost is not discriminative enough to differentiate the true outlier and the suboptimal spots (false positives). However, even under these conditions most of the true outliers are detected by
the proposed algorithm and ultimately the unknown constraint is detected
correctly. This is partly due to the availability of several demonstrations:
the selected inactive points are sufficient to appropriately learn the cost and
the retrieved constraint, which is shown with a red bounding box in the
\figref{fig:man_out}.
Finally, the learned cost and constraints are used to
replicate the skill as shown by the trajectories in \figref{fig:man_gen_traj}.

\section{Conclusion}
\label{sec:conclusions}

In this work, we presented a method to jointly estimate cost and constraints from demonstrations, lifting the reliance of previous methods on known cost function or
their focus on soft constraints alone. 
Moreover, our experiments demonstrate that constraint estimation is sensitive to errors in cost parameters, and a real-world
demonstration involving a Franka Panda manipulator performing a pouring task
validates that the method can be applied in the presence of noisy real
measurements, hinting at the practical usability of the presented method. 

However, one of the main limitations of our method was also shown in the real experiment, where the outlier detection step used in the cost extraction was shown to be sensitive to regions of suboptimality in the demonstration, implying that further work should be devoted to improve the robustness of the detection of inactive segments.\add{ Moreover, the use of more generic cost functions and constraints could better capture complex tasks.}

In conclusion, this work underscores the risk of user-defined cost and the importance of accurately determining the cost function when learning constraints. Our method is a first step in the direction of ensuring the constraints we learn better generalize from the given demonstrations. Moving forward, ensuring reliable learned constraints will be crucial for deploying robots that can safely and efficiently execute learned skills.


\bibliographystyle{IEEEtran}

\end{document}